\documentclass[letterpaper, 10 pt, conference]{ieeeconf}  

\IEEEoverridecommandlockouts                              

\usepackage[pdftex]{graphicx} 
\usepackage{amsmath} 
\usepackage{amssymb}  
\usepackage[caption=false,font=footnotesize]{subfig}
\usepackage{latexsym}
\usepackage{bm}
\usepackage{algorithm}
\usepackage{algorithm}
\usepackage[noend]{algpseudocode}
\renewcommand{\theequation}{\arabic{equation}}
\usepackage{color}
\title{\LARGE \bf
A Framework for On-line Learning of Underwater Vehicles Dynamic Models
}

\author{Bilal Wehbe, Marc Hildebrandt and Frank Kirchner
\thanks{All authors are with  DFKI - Robotic Innovation Center, and the Department of Mathematics and Informatics, University of Bremen, Germany.
       }%
       \thanks{ {\tt\small \{name\}.\{lastname\}@dfki.de}}
}

\begin{document}
\maketitle

\begin{abstract} 
Learning the dynamics of robots from data can help achieve more accurate tracking controllers, or aid their navigation
algorithms. However, when the actual dynamics of the robots change due to external conditions, on-line adaptation of their
models is required to maintain high fidelity performance. In this work, a framework for on-line learning of robot dynamics is
developed to adapt to such changes. The proposed framework employs an incremental support vector regression
method to learn the model sequentially from data streams. In combination with the incremental learning, strategies for
including and forgetting data are developed to obtain better generalization over the whole state space. The framework is 
tested in simulation and real experimental scenarios demonstrating its adaptation capabilities to changes in the 
robot's dynamics.
\end{abstract}

\section{INTRODUCTION}
Accurate modelling of robot dynamics is a critical aspect for most of control algorithms, navigation, path planing and
robot simulation \cite{albiez2015flatfish, hildebrandt2008computerised}. At its core, a model describes the relation between 
the robot's states of motion, the actuation input
and the dynamic forces and torques in play. Typically, models are manually engineered and fined-tuned
to fit a specific robot design or application. This not just the case in underwater systems but also for complex robots like in \cite{lemburg2011AILA}. In future applications we will be able to combine this approach with techniques as presented in \cite{eich2010semantic} and drive underwater vehicles to perform very complex tasks. 

Constructing a model for an underwater system involves computing the inertia of the robot, Coriolis and centripetal 
forces, and external forces such as damping and gravity. Well tuned models would serve the control purpose
given that the robotic system is mechanically stationary and no external disturbances are to be expected. Nevertheless,
a well-engineered but fixed model will naturally have limited usability when the mechanical structure of the 
system changes, or when the operating environment is non-static. This situation arises commonly in
marine robotic applications, where changing a sensor or payload on the robot will result in a completely new
hydrodynamic behaviour, not to mention the environmental disturbances such as bio-fouling or density fluctuations
that can influence the model's integrity.
 
Model learning is a technique that could avoid the manual crafting of the dynamic models, where the dynamic relations
between the robot's actions and states can be directly inferred from the data collected during operation
\cite{nguyen2011model}. In the context of non-stationary dynamics, on-line adaptation is a beneficial perk for any
control or navigation algorithm, since the changes in the system's dynamics can be captured \cite{sigaud2011line}.

\begin{figure}[t!]
\centering
\mbox{
\subfloat[Configuration 1: default configuration]
{\includegraphics[trim=0 0 0 0,clip,width=0.275\textwidth]{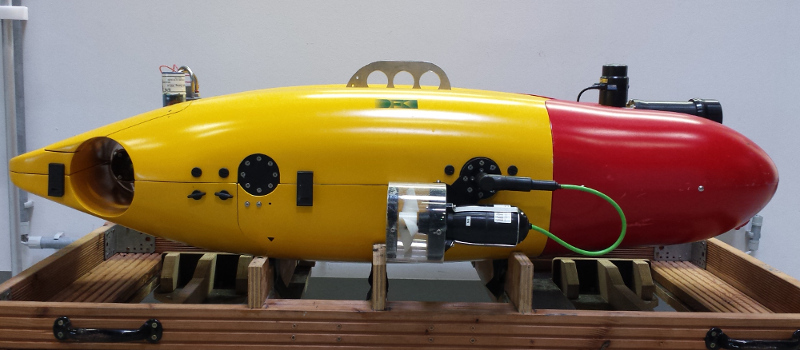}\label{fig:dagon1a}}
\hspace*{0.0cm}
\subfloat[Configuration 1 (perspective view)]
{\includegraphics[trim=0 0 0 0,clip,width=0.17\textwidth]{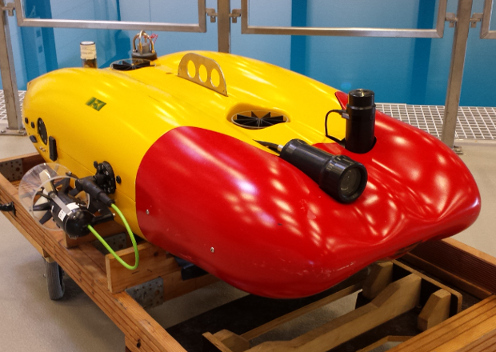}\label{fig:dagon1b}}
}
\mbox{
\subfloat[Configuration 2: with a sonar attached]
{\includegraphics[trim=0 0 0 0,clip,width=0.275\textwidth]{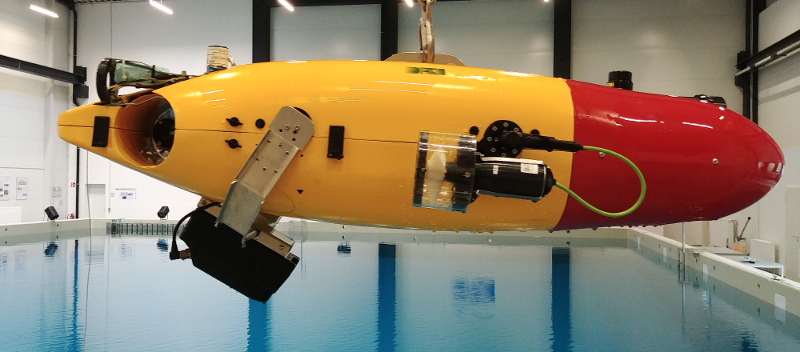}\label{fig:dagon2a}}
\hspace*{0.0cm}
\subfloat[Configuration 2 (rear view)]
{\includegraphics[trim=0 0 0 0,clip,width=0.17\textwidth]{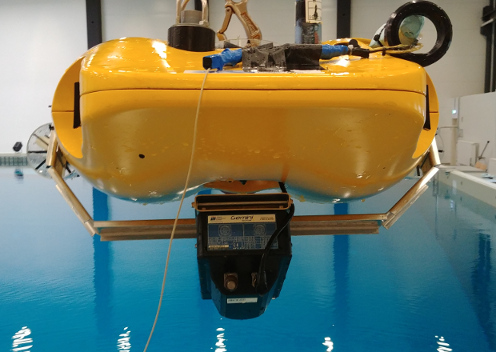}\label{fig:dagon2b}}
}
\mbox{
\subfloat[Configuration 3: with sonar, no front hull.]
{\includegraphics[trim=0 0 0 0,clip,width=0.275\textwidth]{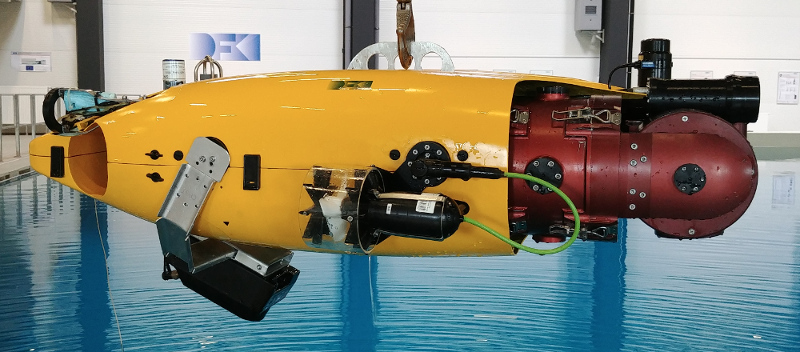}\label{fig:dagon3a}}
\hspace*{0.0cm}
\subfloat[Configuration 3 (front view)]
{\includegraphics[trim=0 0 0 0,clip,width=0.17\textwidth]{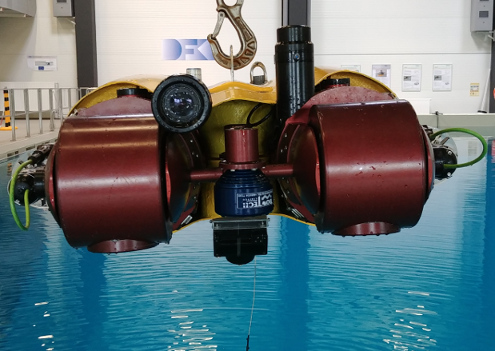}\label{fig:dagon3b}}
}
\caption{The AUV Dagon used as a platform in our experiments. Fig.~\ref{fig:dagon1a} \& \ref{fig:dagon1b} show
the default configuration of the vehicle. Fig.~\ref{fig:dagon2a} \& \ref{fig:dagon2b} show dagon equiped with a sonar.
Fig.~\ref{fig:dagon3a} \& \ref{fig:dagon3b} show dagon equiped with a sonar but with the front hull removed.}
\label{fig:dagon}
\vspace{-0.7cm}
\end{figure}

This paper addresses the problem of on-line learning of robot dynamic models, where we use an autonomous underwater 
vehicle (AUV) named "Dagon" as our test subject. This vehicle was developed in the CUSLAM project
\cite{hildebrandt2010design} and has been extensively modified and used in a number of subsequent research activities 
\cite{hildebrandt2012two, wehbe2017experimental}. This included the addition of a hydrodynamic hull around the pressure 
compartments, as well as the implementation of multiple additional sensor systems. This lead to a number of 
severe changes in the vehicle's hydrodynamic properties, some of which are shown in Fig.~\ref{fig:dagon}. Where, the top 
figures (\ref{fig:dagon1a} \& \ref{fig:dagon1b}) show the current default configuration of Dagon by which it can be used for 
regular survey missions. The middle figures (\ref{fig:dagon2a} \& \ref{fig:dagon2b}) show Dagon when equipped with an 
imaging sonar as payload, whereas the bottom figures (\ref{fig:dagon3a} \& \ref{fig:dagon3b}) show a third configuration 
that represents an accidental partial loss of the vehicle's hydrodynamic hull during operation. 

Building upon our previous findings in \cite{wehbe2017learning, wehbe2017online}, we develop a general
framework for on-line learning of AUV dynamics. Specifically, we use an incremental support vector regression (IncSVR)
algorithm to model the fully coupled non-linear dynamics of the robot. To support the on-line adaptation,
a forgetting strategy is implemented using density estimation technique to regulate pruning of the weights associated to
data samples from different model contexts. Additionally, a node for including new data samples is used to prevent the 
accumulation of redundant data, and an outlier rejection node which is responsible for pruning faulty measurements.
The proposed framework is validated on a dataset collected from a simulation, as well as a real dataset collected
from experimentation with Dagon given the configurations listed above.

\subsection{Related Work}
Robotic model learning has been an active topic of research during the last decade. There has been an extensive work in the
domain of humanoid and manipulator arms dynamics, but not excluding any other domain of robotics. 
A general survey on model learning for robot control can be found in \cite{nguyen2011model}, as well a survey
on on-line regression methods for robot model learning is found in \cite{sigaud2011line}. An infinite mixture of linear
experts approach was used in \cite{jamone2014incremental} to model the dynamic contexts of a manipulator
arm carrying different weights. One drawback of this method is that all training samples need to be stored in the 
memory. Since linear experts model the dynamics only locally, the number of experts increases quickly when modelling
a more complex system. The authors of \cite{jamone2014incremental} show that 60 experts where needed to model
only two contexts of the dynamics. A mixture of Gaussian process (GP) experts with a support vector machine (SVM)
classifier was used to learn different contact models of a manipulator arm in \cite{calandra2015learning}. This method
nevertheless did not show any on-line adaptation rather the models were trained offline. In \cite{mckinnon2017learning},
mixture of GPs was used to model different dynamics of a wheeled robot loaded with
different weights. Benefiting from the Bayesian properties of GPs, a Dirichlet process (DP) was used instead as an
unsupervised classifier. Two drawbacks of this method can be pointed out. Although GPs are commonly used for
modelling robot dynamics, their cubic computational complexity $O(n^3)$ is a major issue when used for on-line
learning \cite{rueckert2017learning}. Secondly, the Dirichlet process classifier is dependent on the density distribution
of the data, i.e. two datasets from the same model but having different density distributions will be classified as two
different models. The authors of \cite{nguyen2009sparse, nguyen2011incremental} proposed an online sparsification 
method by using an independency measure to control the sample pruning. This method nevertheless suffers from high 
computational demand and memory overhead since for a dictionary of saved data points of size $n$, a matrix of size
$(n-1)$ has to be saved for every sample.

In the field of marine robotics, \cite{fagogenis2014improving} used locally weighted projection regression to
compensate the mismatch between the physics based model and the sensors reading of the AUV Nessie. Auto-regressive
networks augmented with a genetic algorithm as a gating network were used to identify the model of a simulated AUV
with variable mass. In a previous work \cite{wehbe2017online}, an on-line adaptation method was proposed to model
the change in the damping forces resulting from a structural change of an AUVs mechanical structure. The algorithm
showed good adaptation capability but was only limited to modelling the damping effect of an AUV model. In this work
we build upon our the results of  \cite{wehbe2017learning, wehbe2017online} to provide a general framework for on-line 
learning of AUV fully coupled nonlinear dynamics, and validating the proposed approach on simulated data as well as real 
robot data.

\section{Online learning of AUV Dynamics}\label{sec:online}
In this section, we introduce the main framework for on-line learning of an AUV dynamic model. First, the dynamics of an
AUV are briefly introduced. Next, we introduce the incremental support vector regression as the core learning algorithm, followed by the strategies for including, forgetting samples and outlier rejection.
\subsection{AUV Dynamics}
The standard dynamics of submersed vehicles are expressed as the combination of Newtonian rigid-body dynamics
and radiation-induced forces and moments \cite{fossen2002marine}. Radiation-induced forces and
moments are expressed as three components: (1) added mass due to the inertia of the surrounding fluid, (2)
potential damping, and (3) restoring forces due to Archimedes. In addition to potential damping, friction due to fluid
viscosity and votrex shedding define the total hydrodynamic damping \cite{fossen2002marine}. Given $\eta$ as the
pose of the body in a fixed reference coordinate frame, and $\nu$ as the velocity in body-fixed coordinates, the full
dynamics of the body can be expressed as
\begin{equation}
 \bm{\dot{\nu}}= M^{-1}\left(\tau + \zeta(\bm{\eta},\bm{\nu},\bm{\tau}) - C(\bm{\nu})\bm{\nu}-d(\bm{\nu})-g(\bm{\eta})\right)\, ,
 \label{dynamic}
\end{equation}
where $M$, $C(\bm{\nu})\bm{\nu}$, $d(\bm{\nu})$, $g(\bm{\eta})$, $\tau$ and $ \zeta(\bm{\eta},\bm{\nu},\bm{\tau})$ are respectively the combined rigid-body and added mass inertia matrix, the Coriolis and centripetal effect, the hydrodynamic damping effect, the restoring forces and moments, the external forces applied to the body and the
unmodelled effects. A compact form of the dynamic model can be written as
\begin{equation}\label{model}
 \bm{\dot{\nu}} = \mathcal{F}(\bm{\eta},\bm{\nu},\bm{\tau})\,.\\
\end{equation}
In this work, we model the thruster dynamics directly with the vehicle model, thus we eq. (2) can be written as
\begin{equation}\label{model2}
 \bm{\dot{\nu}} = \mathcal{F^*}(\bm{\eta},\bm{\nu},\bm{n})\,.\\
\end{equation}
where $\bm{n}$ is a vector representing the rotational velocities of each thruster. 
\subsection{Model Learning with Incremental SVR}
The goal of model learning is to estimate the function $\mathcal{F^*}$ in equation~(\ref{model2}), by having access to its
inputs and outputs. For such purpose, we use a method know as Support vector regression (SVR) \cite{smola2004tutorial}.
One of the advantages of this method is that the model is represented by a smaller subset of the training data known as the 
support vectors $SV$, which we will make use of in the including and forgetting strategies explained in details later. Another 
advantage is the lighter computational cost $O(n^2)$ of this method, as compared to Gaussian process regression.

SVR is a supervised learning method which takes in input-output pairs as training data and learns 
the relation between the input and the output. The goal is thus to fit a function $f(x)=\Phi(x)^T\bm{w}$
onto a training data set $\mathcal{D}=\{(x_i,y_i)|i=1,...,n \}$, where $\Phi(x)$ is a mapping from the input space
onto a higher dimension, and $\bm{w}$ is an associated weight vector. The weight vector can be expressed as a linear 
combination of the input features $\bm{w}=\sum^{i=1}_{n}\omega_i\Phi(x_i)$, thus the regression function can be written as
\begin{equation}
f(x) = \sum^{i=1}_{n}\omega_i\langle \Phi(x_i), \Phi(x)\rangle = \sum^{i=1}_{n}\omega_i \kappa(x_i,x)\, ,
\end{equation}
where $\kappa$ is a kernel function and $\omega_i $ are the linear weights to be estimated.
SVR uses the $\epsilon$-insensitive function defined in \cite{smola2004tutorial} as a loss function, which penalizes the
residual of the predicted output $f(x)$ and its training value $y$ only beyond a margin $\epsilon$. As opposed to
GPR which estimates the weights $\omega_i$ by matrix inversion, SVR solves the problem by using the Lagrangian multipliers 
optimization method. This transforms the optimization into a convex problem as follows (a more detailed explanation can be
found in \cite{smola2004tutorial}) : 

\begin{equation}\label{dual}
 \begin{aligned}
  \min_{\alpha,\beta} & \left\{ \begin{aligned}
                    &  \frac{1}{2}\sum_{i,j=1}^n (\alpha_i-\beta_i)(\alpha_j-\beta_j)\kappa(x_i,x_j) \\
                    & + \epsilon \sum_{i=1}^n (\alpha_i+\beta_i) -\sum_{i=1}^n y_i(\alpha_i-\beta_i)
            \end{aligned} \right. \, ,\\
   s.t.          &  \quad  0 \leq \alpha_i,\beta_i \leq C  \qquad \forall i : 1 \leq i \leq n \, , \\
                  &  \quad \sum_{i=1}^n (\alpha_i-\beta_i)=0.
 \end{aligned}
\end{equation}
Where $\alpha_i,\,\beta$ are the Lagrangian multipliers, $C$ is a constant that compromises between having a more
generalizing model with low weights or having too large deviations, and $\kappa(x,z)$ the kernel function. To solve
(\ref{dual}), the sequential minimal optimization method, as implemented in \cite{CC01a} was used, resulting in
the final regression function
\begin{equation}
 f(x)=\sum_{i=1}^m(\alpha_i-\beta_i)\kappa(x_i,x)\, .
\end{equation}
As kernel we use the squared exponential defined as 
\begin{equation}
\kappa(x,x') = \text{exp}\left(-(x-x')^T S^{-1} (x-x')\right)
\end{equation}
where $S=\Sigma/\gamma$ is a matrix proportional to the covariance of the training data.

One advantage of SVRs is the $\epsilon$-insensitive loss function where all samples with a residual below the threshold
$\epsilon$ are assigned a zero weight; thus, the set of samples left are used for producing predictions. These
samples are called the set of support vectors $SV$. Another beneficial aspect of solving for the weights as an optimization
problem is the quadratic computational complexity $O(n^2)$ in contrast to matrix inversion which is of cubic complexity
$O(n^3)$.

However, for on-line learning continuous updates of the estimated weights is required as the data arrives sequentially.
Starting from a set of support vectors with their corresponding weights, as a new sample arrives it is added on top of
the existing samples and thus the new set is passed to the optimization algorithm. The weights of the already existing
samples are kept as a starting point for the next optimization step which results in a much faster conversion since we
start from a more optimal solution. Nevertheless, having a stream of continuously arriving data will lead to an
unlimited growth of the set of support samples causing the memory to increase boundlessly. To obey the memory and computational
constraints, we limit the set of support samples to a fixed buffer. Therefore as new samples get in, older samples has
to be removed to keep a fixed size of the buffer. The straight-forward way is to first remove the oldest samples in the
buffer, or what is commonly known as first-in-first-out (FIFO), since assumingly they are the most outdated samples.
However, this approach could lead to a dangerous situation where the model will lose information about certain
regions of the model space, if for a certain reason the new coming samples get concentrated in a local region.
For example, if the mission requirements demands the robot to operate in low speeds for a prolonged time, the
information about the dynamics in higher speeds can be lost over time.
To cope with such shortcoming we propose the following strategies to control the adding and removal of data samples
form the SVR buffer.

\subsubsection{Forgetting strategy}
Given a limited buffer size, the main idea of the forgetting strategy is to keep a balanced global distribution of support 
vectors over the model's sample space , as well as keeping the support samples inside the buffer up-to-date. However,
there must be a trade-off between the density distribution of a sample inside the buffer and its queue time. The higher the
density and the older a sample is, the more likely this sample will be removed. We define thus the following metric
\begin{equation}\label{forget}
\phi = \frac{d}{\sqrt{t_s} +k}\, ,
\end{equation}
where $d$, $t_s$ and $k$ are respectively the density of a sample, the time stamp when the sample was recorded,
and a constant weighting the importance of the age of the sample over its density. We use the square root of the
timestamp to avoid very high values which would lead the forgetting score $\phi$ to approach zero. Eventually, the 
samples with the highest scores will be removed from the buffer until the maximum allowed size is reached.

To estimate the density of the samples we use multivariate kernel density estimation (KDE), where we average the correlation 
of a sample with respect to its neighbouring observations. The kernel function a measure of the correlation between two 
samples, therefore a higher weight is given if the samples are closer to each other whereas
the weight vanishes as the samples get more distant. The general equation of a KDE can be written as
\begin{equation}\label{density}
d(x) = \frac{1}{n} \sum_{i=1}^{n} \frac{1}{det(H)} \kappa\left( H^{-1} (x-x_i)\right) \, ,
\end{equation}
where $H$ is a nonsingular bandwidth matrix which need to be tuned carefully. A cross-validation 
optimization is quite costly with more variables in the input features vector. To avoid computational power, we can 
reduce the cross-validation optimization to tuning only one parameter by choosing a bandwidth matrix that is 
proportional to the covariance of the input data \cite{hardle2012nonparametric}.

\subsubsection{Including strategy}
As the stream of data is fed into the SVR learner, every sample is passed first through a filtering gate to
determine if it will be used by the learner or discarded from the training step. The main motivation behind this approach
is to save memory and computational power and as well prevent redundancy in the set of support vectors learned by the
model. In practice, new samples that fall in a very close proximity of already existing support samples, or the residual
of the target and the prediction is less than the threshold $\epsilon$ will not have any significant influence on the 
regression function. Additionally we make benefit from the kernel function to measure the proximity of a newly arriving
sample with respect to the support vectors. Thus, a new sample $(x_i,y_i)$ is discarded from training if for any support
vector $(x_{sv},y_{sv})$  one of the following is true
\begin{equation}\label{eq:inc_1}
\kappa(x_i,x_{sv}) > \xi  \quad  \text{and}  \quad |f(x_i)-y_i|<\epsilon
\end{equation}
\begin{equation}\label{eq:inc_2}
\kappa(x_i,x_{sv}) > a \quad  \text{and}  \quad  |y_i - y_{sv}|<b
\end{equation}
where $\xi$, $a$ and $b$ are constant thresholds that can be selected through cross validation.

\begin{figure}[t]
\centering
\vspace{0.3cm}
\includegraphics[trim=0 0 0 0,clip,width=0.48\textwidth]{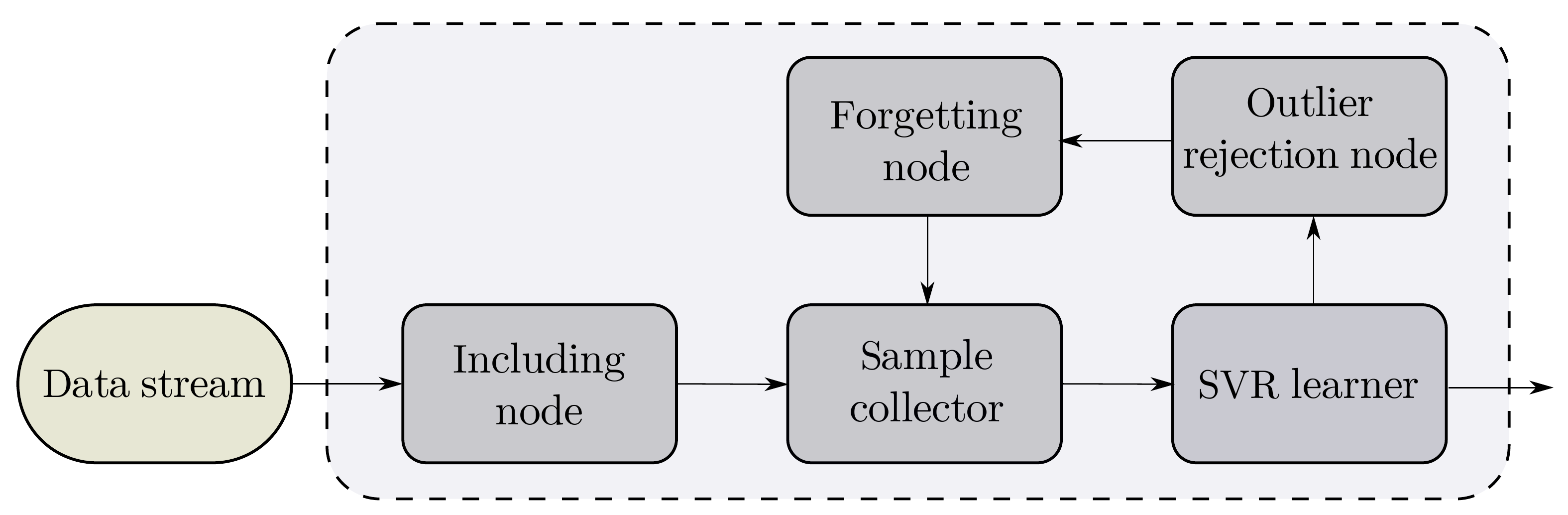}
\caption{Flow diagram of the on-line training framework.}
\label{fig:framework}
\end{figure}

\subsubsection{Outlier rejection}
The last node of the framework is to deal with outliers in the training data due to faulty sensor measurements. Although
most outliers will be filtered out in the including node due to equation (\ref{eq:inc_2}), a sample that is not lying in the 
vicinity of any support sample would still be considered as a novelty and admitted in for training. Therefore, the outlier filter is 
applied on the set of support samples after each training step. To satisfy on-line training restraints, we keep a computationally 
low-cost approach for filtering outliers. We compute the residue of each support sample and its corresponding prediction
\begin{equation}
 \text{res}_{sv}=f(x_{sv})-y_{sv}.
\end{equation}
The interquarile range (IQR) of the residuals are then computed and all samples with their corresponding residuals
falling outside side of the range of 
\begin{equation}\label{IQR}
\text{IQR} = [q1-1.5(q3-q1),q3+1.5(q3-q1)] 
\end{equation}
are flagged as outliers. Here, $q1$ and $q3$
represent the limits of the first and last quartile \cite{rousseeuw1993alternatives}.

Finally, the overall framework can be viewed as five main nodes as shown in Fig.~\ref{fig:framework}. Data samples
will be received from the stream into the including node first, then passed into the sample collector where they are
merged with the processed buffer of support samples from the previous step. The stack of new training samples and older
support samples are then fed into the training node. The resulting new set of support vectors are then processed by the 
outlier rejection and forgetting nodes and made ready for the next iteration. A pseudocode of the overall framework can be 
found in Algorithm~\ref{alg1}. We note here that although our framework uses SVR, which unlike GPR, does not estimate a
true confidence interval on prediction, using the kernel density estimation can provide a measure of uncertainty 
of a prediction.
\begin{algorithm}[t!] 
\caption{On-line Learning Framework} 
\label{alg1} 
\begin{algorithmic} 
    \State {\bf Input:} new data sample $(x_i,y_i)$
    \State {\bf Including:}
    \For {$(x_{sv},y_{sv})$ in $SV$}  
    \If{eq. (\ref{eq:inc_1}) or (\ref{eq:inc_2}) are true}
        \State delete $(x_i,y_i)$
    \EndIf 
    \EndFor 
    \State {\bf Sample collector:}
    \If {$(x_i,y_i)$ is not empty}
        \State concatenate $(x_i,y_i)$ and $SV$
    \EndIf
    \State {\bf Training:}
    \State solve eq.~(\ref{dual}): compute $\alpha_i$ and $\beta_i$
    \State {\bf Outlier rejection:}
    \For {$(x_{sv},y_{sv})$ in $SV$}
        \State compute $ \text{res}_{sv}=f(x_{sv})-y_{sv}$
    \EndFor
    \State calculate IQR using eq.~(\ref{IQR})
    \State delete $(x_{sv},y_{sv})$ with the corresponding res is outside IQR
    \State {\bf Forgetting:}
    \For {$(x_{sv},y_{sv})$ in $SV$}
        \State compute $\phi$ using eq.~(\ref{forget}) and (\ref{density})
    \EndFor
    \While{size of $SV >$ buffer size}
        \State delete $(x_{sv},y_{sv})$ with the highest $\phi$
    \EndWhile
\end{algorithmic}
\end{algorithm}
\section{Evaluation and results}
In this section, we present the evaluation procedure of our approach, where we test its performance on two
datasets, one acquired from simulation and one from an actual experimental trial with the robot.
Dagon is a hovering type AUV equipped with two vertical thrusters for depth and pitch stabilization (roll is passively
stable), and three lateral thrusters that are used differentially to control the vehicle in the surge, sway and yaw
directions. Stabilizing the vehicle in depth and pitch will help us to reduce the dimensionality of the problem, where the
regression problem is then formulated as a mapping between an input feature vector composed of the vehicle's
surge, sway and yaw velocities and the lateral thrusters' rotational speed, and an output target vector defined as the
the vehicle's  acceleration in surge sway and yaw.

\subsection{Data Acquisition}
\subsubsection{Simulation Setup}
As a first evaluation step of our on-line learning framework, we design a simulation environment for Dagon that allows
us to easily induce changes into the dynamics. For the dynamics simulation we use the underwater dynamics 
plugin\footnote{github.com$/$rock-gazebo$/$simulation-gazebo$\_$underwater} \cite{britto2017improvements}
for the Gazebo simulator \cite{koenig2004design}. A dataset is collected for three different configurations of the vehicle's 
dynamics, where the first simulates the default configuration of the robot, the second represents a damage in one of its 
thrusters, and the third represents a change in the damping effect due to a mechanical change in the vehicle's structure.
Actuation inputs are given to the later thrusters in the form of a sinusoidal signal of varying frequencies in order to cover a 
bigger range of the model's state space. Every thruster is given a different sine wave with a period that changes randomly 
between 20 and 70 seconds. We sample the simulation at a frequency of 1 Hz, generating a dataset of 30,000 samples in 
total, divided equally between the three mentioned configurations.

\subsubsection{Experimental Setup}
A set of experiments were carried out with Dagon in a salty water basin with a static water volume and no induced
currents. The vehicle was controlled in a similar fashion as in the simulation, where we stabilize pitch and depth and let 
it run freely in a horizontal plane by using its lateral thruster. Dagon is equipped with a number of navigation sensors 
from which we will use a doppler velocity log (DVL) to measure the linear velocities and a fiber-optics-gyroscope to
measure the angular velocities. All thrusters are also equipped with Hall effect sensors to measure their rotational 
speeds. Similar to the simulation runs, the thrusters receive a separate sinusoidal command with randomly shifting
periods. A separate trial was conducted for each of the configurations shown in Fig.~\ref{fig:dagon}, resulting 
in a dataset with 31300 samples, split as 11567, 10050 and 9683 samples respectively for configurations 1, 2 and 3.

\subsection{Offline Learning}
\begin{figure}[t!]
\vspace{0.3cm}
\centering
\includegraphics[trim=30 5 39 5,clip,width=0.48\textwidth]{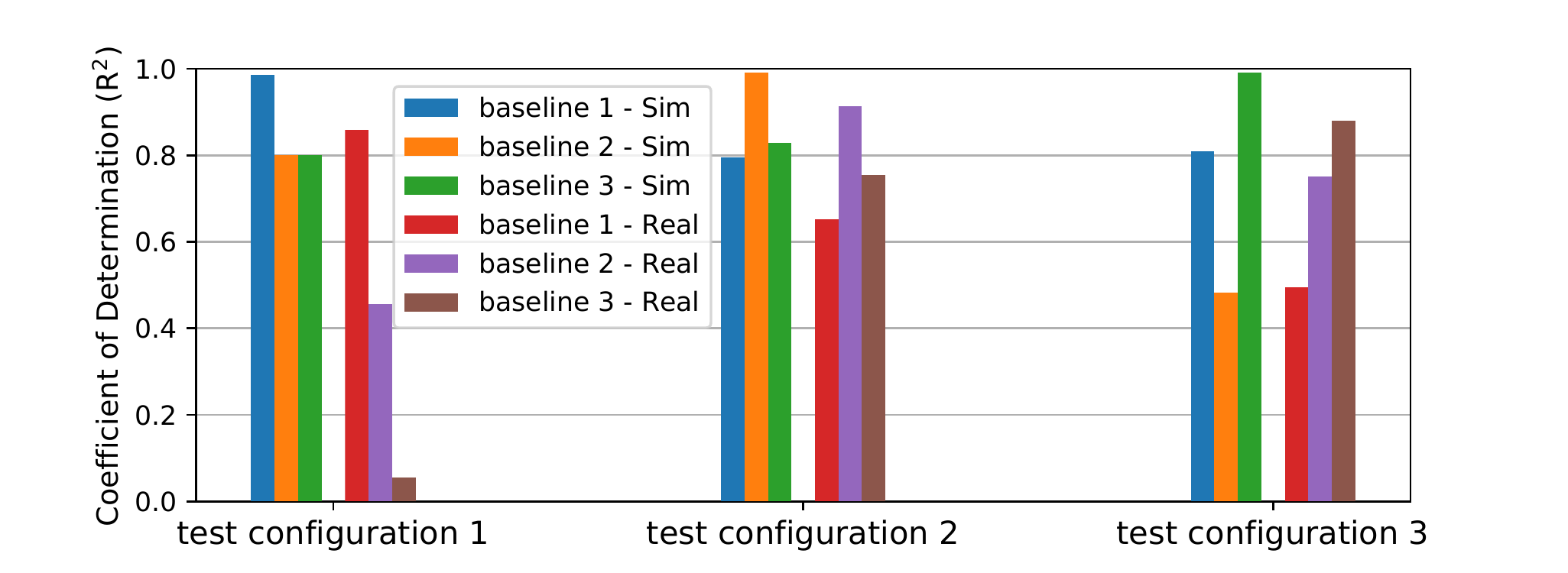}
\caption{Validation results of the offline trained models. Every model is trained with data from one configuration at a time, 
and then tested against the validation data of all three configurations. The validation results of each model shows a high score
when trained and tested on data from a single configuration whereas a lower score is observed when testing with other 
configurations.}
\label{fig:offline}
 \vspace{-0.25cm}
\end{figure}
For each of the two test scenarios; simulation and real data;  80$\%$ of the data is used for training and 20$\%$ for 
validation. The splits are done in a stratified manner, which means for each different configuration, 20$\%$ of the data is left 
out for validation and the rest is used for training. As a goodness-of-fit scoring metric, we use the coefficient of 
determination:
\begin{equation}
 R^2 = 1- \frac{\sum_{i=1}^{n} (y^{predicted}_i-y^{true}_i)^2}{\sum_{i=1}^{n} (\bar{y}-y^{true}_i)^2},\\
\end{equation}
where $\bar{y}$ is the mean of  $y^{true}$.
For each of the two datasets, we train three separate models as baselines for our evaluation. This means, a separate SVR is 
trained with the data corresponding to each configuration. Thus, the notation \emph{"Baseline 1 - Real"} means a model is
trained with the data from configuration 1 of the real dataset, \emph{"Baseline 2 - Sim"} would indicate a model is trained
with the second configuration of the simulation data, and so forth. Each baseline model is a supervised SVR by itself that has 
been cross-validated with the validation data to ensure the highest performance that could achieved. In Fig.~\ref{fig:offline} 
we report the validation scores of all six models tested with the validation sets of every configuration separately. The blue, 
orange and green bars represent the scores of the baselines trained with configurations 1, 2 and 3 of the simulation data 
respectively, whereas the red, purple and brown bars correspond to the baselines of the real data. It can be observed in both 
scenarios, that every model achieves a high validation score when it is trained and tested on data from the same 
configuration, whereas lower scores are reported when tested on data from other configurations. These results are not very
surprising as they demonstrate that any static model trained with one configuration of dynamics cannot describe accurately
other configurations, which emphasizes the need of on-line adaptation.
\begin{figure*}[h!]
\centering
\vspace{0.3cm}
\includegraphics[trim=130 0 150 30,clip,width=0.99\textwidth]{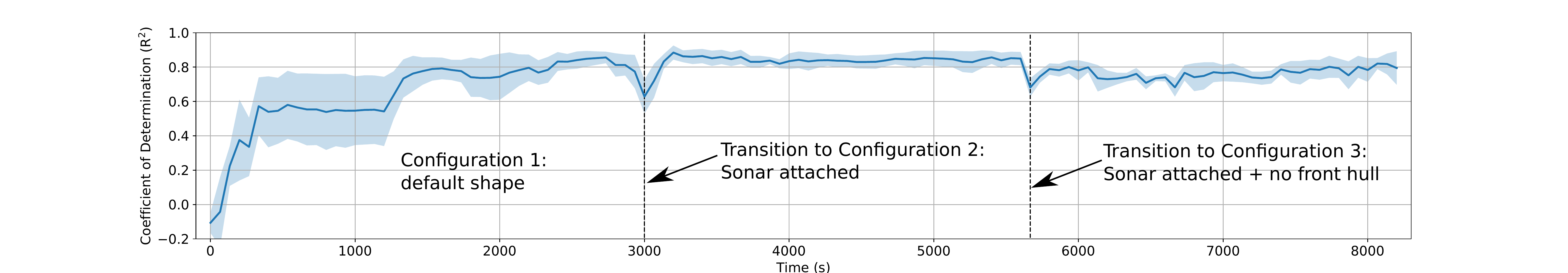}
\caption{This figure shows the evolution of the performance score $(R^2)$ on the validation data as the training data is
fed in to the estimator. The interval (0-3000s) represents the default configuration of Dagon, where one can notice the 
prediction on the  validation data gets more and more accurate as the data flows in. The second interval (3000-5700s) 
corresponds to the data from the second configuration. Here the validation data is changed also the second 
configuration, which explains the drop in the performance. The accuracy then increases gradually as the 
model adapts to the newly arriving data, reaching a score equivalent to the offline baseline. The last interval (5700-8300s) 
shows the adaptation to the third configuration of the robot.}
\label{fig:validation}
\end{figure*}

\begin{figure*}[t!]
\centering
\subfloat [Simulation data]{
\includegraphics[trim=10 10 0 5,clip,width=0.48\textwidth]{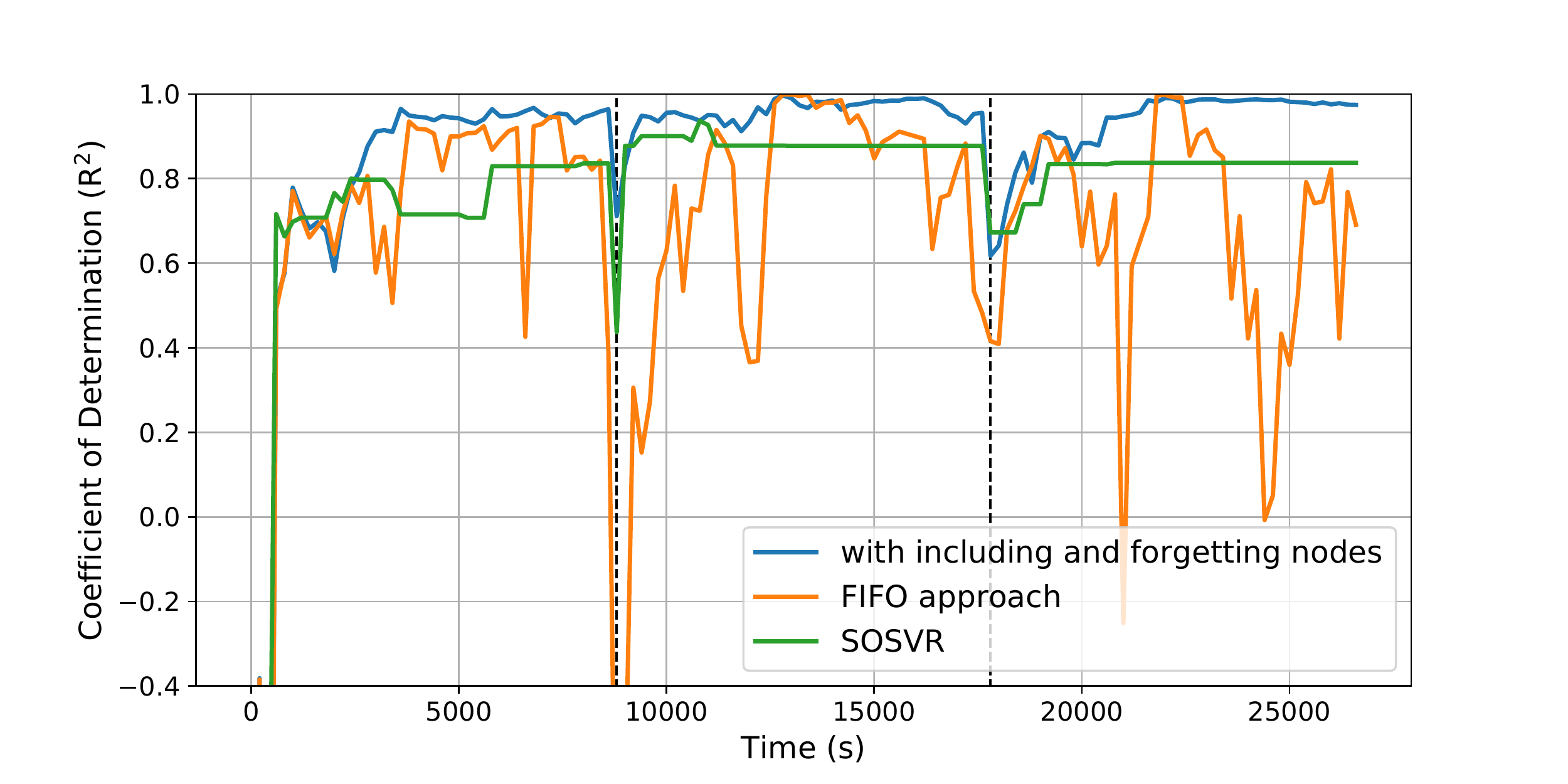}}
\subfloat[Real data]{
\includegraphics[trim=0 10 10 5,clip,width=0.48\textwidth]{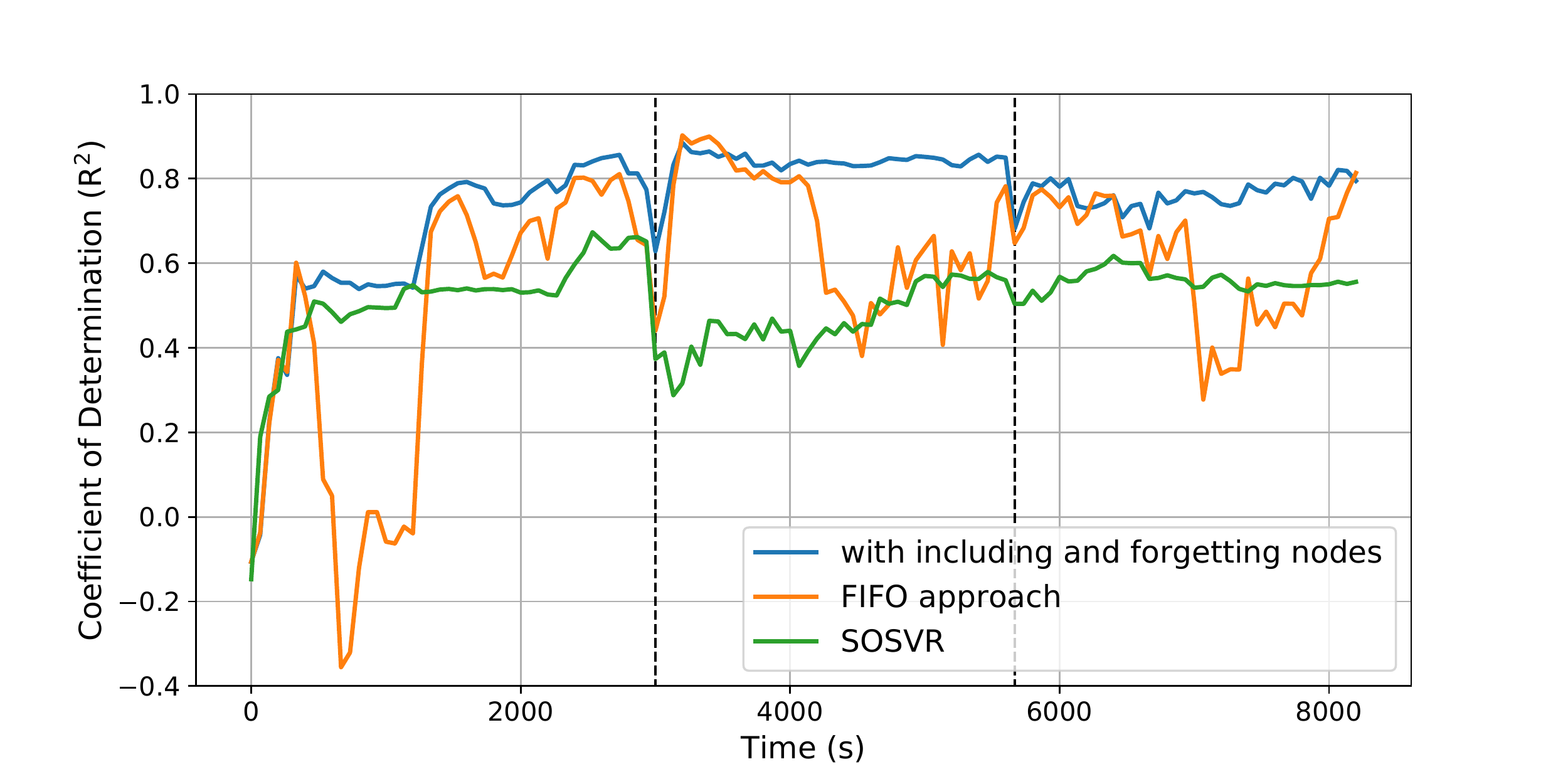}}
\caption{Comparison between the proposed framework, SOSVR and the FIFO approach. (a) shows the results from the 
simulation data whereas (b) shows the results of the real data. The blue line in both cases represents the $R^2$ score of the 
proposed framework the including and forgetting strategies, whereas the orange line shows the performance with a FIFO
approach, and the SOSVR in green. In both scenarios the  FIFO approach reports an unstable performance since only the 
oldest samples are removed from the buffer without taking into account the distribution of support samples across the 
sample space.}
\label{fig:comparison}
\end{figure*}
\subsection{On-line Learning with Changing Dynamics}\label{sec:training}
In this section we demonstrate the capability of the proposed algorithm of on-line learning and adapting to the changes in 
dynamics. We use the same splits of the data as of the offline scenario, but this time the data is provided sequentially to the
learner. The evaluation scheme used is described as the following. (1) We continuously evaluate the performance by testing 
on the validation data after every training step, as the training data stream is fed in. (2) We evaluate always on the validation 
data that corresponds to the training data seen by the learner, i.e., if training data from the first configuration is observed by 
the learner, then evaluation is done on the validation data from the first configuration as well.
\begin{table}
\vspace{0.4cm}
 \caption{Results of hyperparameter optimization}
 \label{tab:hyperparam}
 \centering
\begin{tabular}{c | c | c | c }
\hline
Hyperparameter &  Surge &  Sway & Yaw\\
\hline
 $\epsilon$    & 0.1 & 0.001 & 0.1   \\
 $C$             & 10   & 10 & 10\\
 $\gamma$  & 100 & 40 & 20\\
 buffer size  & 900 & 900 & 900\\
 $k$             & 10 & 10 & 1\\
 $a$            & 0.99& 0.99 & 0.99\\
 $b$            & 10$^{-2}$ & 10$^{-2}$ & 10$^{-2}$\\
 $\xi$          & 0.99 & 0.99 & 0.99\\
\hline
\end{tabular}
\vspace{-0.5cm}
\end{table}
First, the hyper-parameters for the SVR as well as the additional parameters due to the including and forgetting nodes were 
optimized using only the data from the first configuration, and fixed for the rest.  The values
of the hyper-parameters chosen via cross-validation are reported in TABLE~\ref{tab:hyperparam}.

\begin{figure}[t!]
\vspace{0.3cm}
\centering
\includegraphics[trim=30 5 40 5,clip,width=0.5\textwidth]{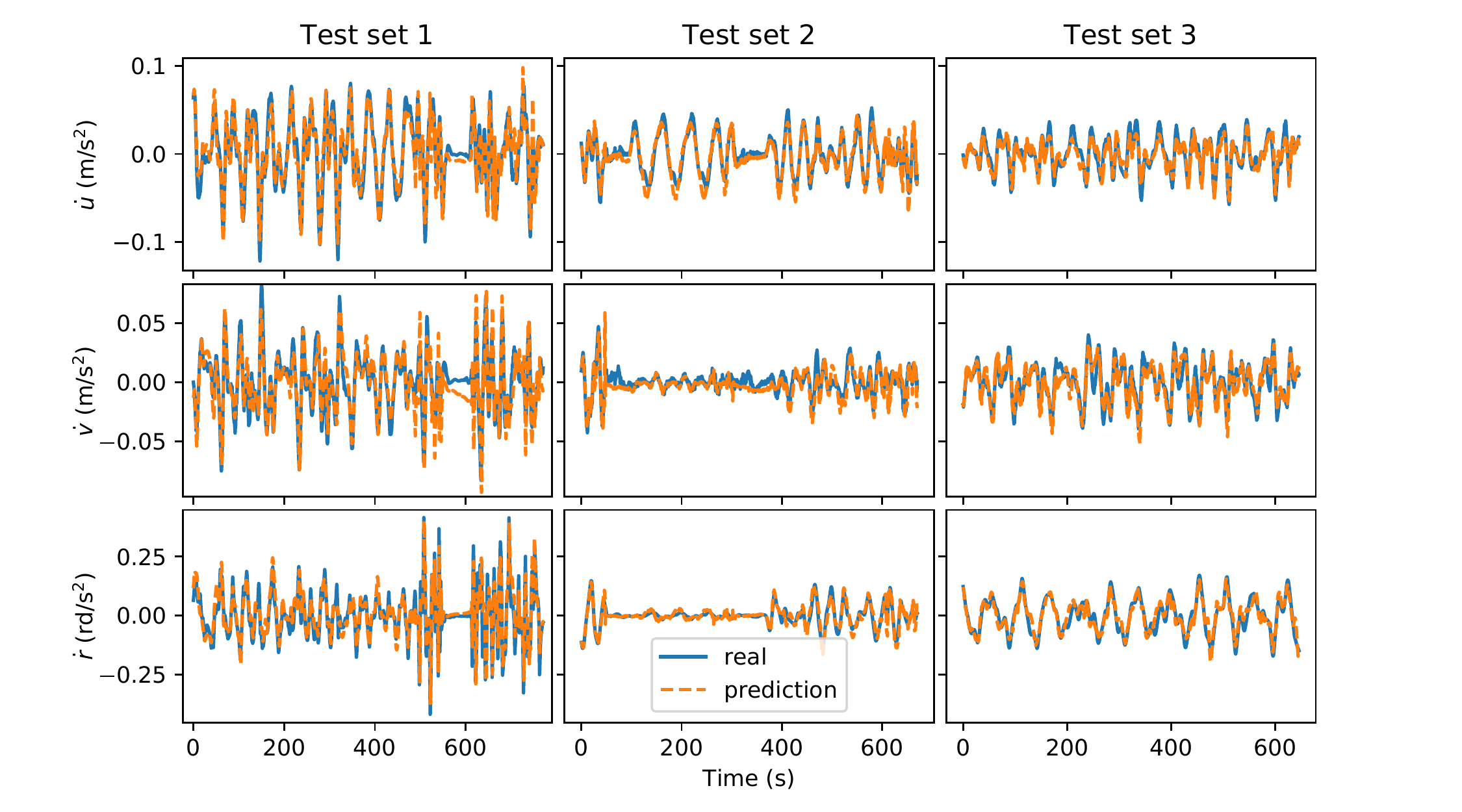}
\caption{Prediction results computed at the end of every training set just before switching to a new configuration. Note that all test sets are separate data that the online model did not see at training time.}
\label{fig:prediction}
 \vspace{-0.25cm}
\end{figure}

In Fig.~\ref{fig:validation}, the propagation of the $R^2$ score of the real data is plotted, as the training takes place. This score results from testing against the validation data after every training step. The training is started from scratch without any prior knowledge of the model. We start by feeding in the training data from the first configuration of the robot, at this point the validation data correspond to the first configuration as well. At the early stages, a very low performance can be observed since the model has not gained enough information about the full dynamics. As time passes by, the learner experiences more training data, which shows a gradual increase in the performance until it reaches a comparable value to the offline baseline (\emph{"Baseline 1 - Real"}). At this point, a general description of the dynamics of the first configuration is learned. At around 3000 seconds, a transition to the second configuration of the robot starts. From this point on, the validation is switched to the second configuration as well, where a drop in the performance can be noticed since the memory of the learner is still populated with information about the old configuration. As time progresses, the evaluation score increases again as the model adapts to the new configuration. A similar behaviour can also be observed from 5700 seconds on, where the data from the third configuration flows in. It can be noticed that the proposed on-line method adapts well to new configurations of the dynamic, as well as being able to reach, in every configuration, a performance comparable to the corresponding offline baselines. After every training set is finished, the prediction results over the corresponding test set are shown in Fig.~\ref{fig:prediction}.

\subsection{Discussion}
In this section we demonstrate the necessity of the including and forgetting strategies as we compare the performance of the 
proposed framework with the FIFO approach as well as the sparse-online-SVR (SOSVR) presented in \cite{nguyen2009sparse}. For the FIFO model, we use the same incremental SVR approach with a fixed buffer, but with removing the oldest samples first as new samples are included. A grid search is used to optimize the hyperparameters of the SOSVR.
Fig.~\ref{fig:comparison} shows a comparison between all three methods for the simulation and the real test scenarios. The scores of the method presented in this paper are showed in blue whereas the FIFO method is plotted in orange and the SOSVR in green. In the early stages of training, it can be observed that all methods report similar validation scores. As the training continues, our approach shows a consistent performance in both cases, where the performance converges to a stable and accurate state. On the other hand, as the buffer of the FIFO method gets fully occupied, removing only the oldest samples in the buffer results in an unstable and jittery performance. The fluctuation of the validation score of the FIFO method can be observed throughout the whole training process. This behaviour can be interpreted by the uncontrolled pruning of data, which results in loss of important information in some regions of the model's state space. Alternatively, incorporating the density of the support samples into the pruning procedure, helps keeping a balanced distribution, yet up-to-date samples in the learning buffer. On the other hand, the SOSVR method shows consistent adaptation with changing dynamics but a significant lower accuracy compared to the method proposed in this work.
\section{CONCLUSIONS}
In this work we presented a framework for learning on-line the model of a robot and adapting to changes in its dynamics.
An including and forgetting strategies were developed to control the pruning of old data data, without losing information 
about the global state space of the model. The proposed framework was validated in two test scenarios, a simulation
and real experimental data from an AUV with different configurations, which showed its adaptation capabilities to the new
dynamics.

As future work, we aim extend this method to include a database where the learned models can be stored and reused by the 
robot if a similar situation is encountered again.





\section*{ACKNOWLEDGMENT}
This work was supported by the Mare-IT (grant No. O1lS17029A) and  EurEx-SiLaNa (grant No. 50NA1704) projects which are funded by the German Federal Ministry of Economics and Technology (BMWi).


\bibliographystyle{IEEEtran}
\bibliography{IEEEabrv,library}

\end{document}